\documentclass[conference]{IEEEtran}

\usepackage[square,sort,comma,numbers]{natbib}
\usepackage{graphicx}
\usepackage{amssymb}
\usepackage{algorithm}
\usepackage{algpseudocode}
\usepackage{amsmath}

\DeclareMathOperator*{\argmin}{arg\,min}
\setcitestyle{square}
\title{\LARGE \bf
Collision Avoidance Robotics Via Meta-Learning (CARML)
}
\author{Aravind Mahadevan \\
        armahade@eng.ucsd.edu \\
        University of California, San Diego
\and 
Abhiram Iyer \\
abiyer@ucsd.edu \\
University of California, San Diego
}
\begin{document}
\maketitle
\thispagestyle{plain}
\pagestyle{plain}


\begin{abstract}
    This paper presents an approach to exploring a multi-objective reinforcement learning problem with Model-Agnostic Meta-Learning. The environment we used consists of a 2D vehicle equipped with a LIDAR sensor. The goal of the environment is to reach some pre-determined target location but also effectively avoid any obstacles it may find along its path. We also compare this approach against a baseline TD3 solution that attempts to solve the same problem.
\end{abstract}

\section{INTRODUCTION}
Today, most deep reinforcement learning techniques require models to be trained on a large number of training samples. In contrast, Model-Agnostic Meta-Learning (MAML) proposed by Finn et. al. \cite{maml} requires  fewer training samples because it tries to train models that can quickly adapt to new tasks. Specifically, MAML is shown to effectively train simulations of an ant or cheetah as they learn to walk across an obstacle-free environment. In this paper, we explore whether MAML can be applied to train agents on multi-objective based navigation tasks. We use MAML to train agents (vehicles equipped with a LIDAR sensor) on the Brain Corp robot motion planning  environment \cite{bcgym} to avoid randomized obstacles and reach a target destination without collision.

An example of this environment can be seen below, where the vehicle is highlighted in green, the obstacle in grey, and the goal location in red:

\begin{figure}[h]
    \centering
    \includegraphics[scale=0.6]{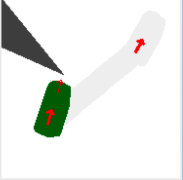}
    \caption{Brain Corp robot motion planning environment}  
    \label{fig:environment}
\end{figure}

\section{RELATED WORKS}
Meta-learning has become quite popular in the last few years due its ability to quickly adapt to new tasks with few training examples. Some of the earliest work on meta-learning was done by Schmidhuber et al. in \cite{schmidhuber-1987}, \cite{schmidhuber-1992}, and \cite{schmidhuber-1993} where he investigated networks that are able to tune its own weights. Today, there are three commonly used frameworks for meta-reinforcement learning algorithms: optimization-based, metric-based, and model-based learning. Since our work primarily deals with MAML, we will mainly be focusing on optimization-based meta learning techniques in this section. 

In optimization-based meta learning, deep learning models are trained with backpropagation with the goal of converging given only a few training examples. One such example of this approach is by Ravi and Larochelle’s work in \cite{ravi-larochelle}. Inspired by the work done by Andrychowicz et al. in \cite{Andrychowicz-work}, they modeled an LSTM as a meta-learner, which helped to train another neural network "learner" classifier using a few-shot framework. Unlike common deep learning optimizers such as Momentum, ADAM, and Adagrad, this method is able to train a model that can adapt to a different data set without needing to retrain the entire network. Unlike Ravi and Larochelle’s work in \cite{ravi-larochelle}, MAML is a much general procedure that only requires that the model be trained using gradient descent. The one downside of this method is the need to compute the Hessian, which affects the overall speed of this algorithm. The authors also noted that the model still performs well when taking a first order approximation of the gradient in the meta-optimization step. Similar to MAML, Nichol et al. proposed the Reptile method in \cite{Reptile-paper} which similarly uses the first order approximation of MAML. However, its optimization step is more similar to joint training and does not require a training-test split. Furthermore, the authors in Reptile also give empirical analysis to show that the gradient will move the parameters of the model closer to all the optimal manifolds of all tasks used. 

\section{BACKGROUND}
In this section, we will discuss the two algorithms we used to solve our task: MAML and TD3. We will first present the MAML algorithm and use similar nomenclature to that in \cite{maml}. For more details on this algorithm, we suggest referring to \cite{maml}. We will then present the baseline TD3 algorithm which we used to compare the performance of our agent trained with MAML. Again, for more details, we suggest referring to \cite{fujimoto2018addressing} for more details.  

\subsection{Model Agnostic Meta Learning}

We first, consider a model $f_{\theta}$ that is parameterized by weights $\theta$ which represents the policy. We have a task distribution $p$($\mathcal{T}$) for which the model is trained on and later adapted to.  Each task $\mathcal{T}$ = $\{\mathcal{L}\{\mathbf{x_1, a_1, ..., x_H, a_H}\}, \textit{q}(\mathbf{x_1}),\textit{q}(\mathbf{x_{t+1}}|\mathbf{x_{t}},\mathbf{a_{t}}), \textit{H} \}$ has a loss function $\mathcal{L}$, an initial observation distribution, $\textit{q}(\mathbf{x_1})$, a transition distribution $\textit{q}(\mathbf{x_{t+1}}|\mathbf{x_{t}},\mathbf{a_{t}})$, and an episode length of $\textit{H}$.  The loss function  $\mathcal{L}\{\mathbf{x_1, a_1, ..., x_H, a_H}\}$  is task specific and, in general, is a negative reward based loss for reinforcement learning problems. 
MAML has two phases: meta-learning and adaptation. During meta-learning, $f_{\theta}$ is trained to obtain a good weight initialization so that $f_{\theta}$ is able to learn a new task quickly during the adaptation phase. The illustration in Figure \ref{fig:mamlIllustration} demonstrates how the two phases are related.
 During the meta-learning phase, we first sample a batch of tasks $\mathcal{T}_{i} \sim p$($\mathcal{T}$). One thing to note is that tasks used for testing are excluded in the meta-learning phase.  
 For each task, we obtain $\textit{K}$ trajectories of maximum length $H$. From these $\textit{K}$ trajectories, we obtain adapted parameters $\theta_{i}'$ for a task $\mathcal{T}_{i}$. To obtain the adapted parameters $\theta_{i}'$, the authors used a policy gradient update with baseline proposed by \cite{gae} and used one or more updates to obtain the adapted parameters $\theta_{i}'$.

After computing the the adapted parameters for a task, $\theta_{i}'$ is fitted to the model to get $f_{\theta_{i}'}$. $f_{\theta_{i}'}$ is then used to generate a new set of trajectories which will be used to update $\theta$. Once the adapted parameters are computed across all tasks in the batch, $\theta$ is updated. $\theta$ is optimized so that the loss of $f_{\theta_{i}'}$ is minimized for task $\mathcal{T}_{i}$ and for all tasks in the current batch sampled from $p(\mathcal{T})$. Specifically, we minimize the following meta-objective function: 
 \begin{equation}
 \min_{\theta} \sum_{\mathcal{T}_{i} \sim p(\mathcal{T})} \mathcal{L}_{\mathcal{T}_i}(f_{\theta_{i}'}) = \sum_{\mathcal{T}_{i} \sim p(\mathcal{T})} \mathcal{L}_{\mathcal{T}_i}(f_{\theta - \alpha\nabla_{\theta}\mathcal{L}_{\mathcal{T}_i}(f_\theta)})
\end{equation}

To minimize the objective function, the model parameters, $\theta$, are updated in the following manner: 
\begin{equation}
    \theta = \theta - \beta\sum_{\mathcal{T}_{i} \sim p(\mathcal{T})} \mathcal{L}_{\mathcal{T}_i}(f_{\theta_{i}'})
\end{equation}
To update $\theta$, the authors used trust-region policy optimization  as the meta-optimizer for the reinforcement learning tasks. \cite{trpo} Algorithm \ref{alg:maml_alg} presents the full MAML algorithm for reinforcement learning which uses the following general loss function for reinforcement learning tasks: 
\begin{figure}[t!]
    \centering
    \includegraphics[scale=0.35]{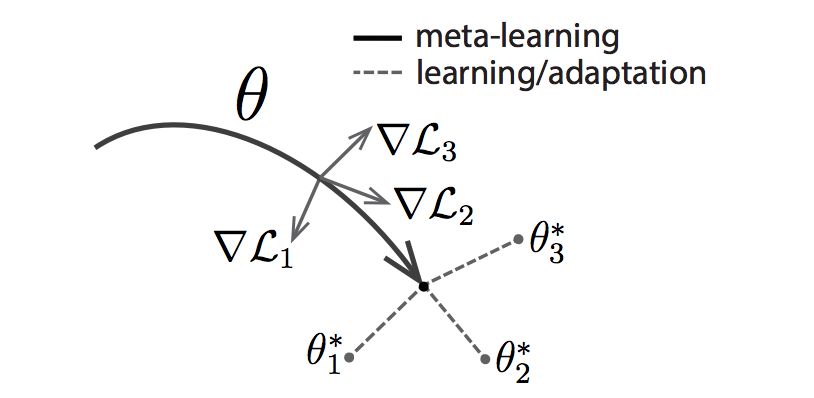}
    \caption{Two phases of meta learning. During meta-learning, the model parameters are trained in such a way that model can adapt quickly to learning new tasks. Figure found in \cite{maml}}   
    \label{fig:mamlIllustration}
\end{figure}

\begin{equation}
  \mathcal{L}_{\mathcal{T}_i}(f_\theta)= -\mathbb{E}_{\mathbf{x}_t, \mathbf{a}_t \sim f_\theta ,\textit{q}(\mathbf{x}_t)}[\sum_{t=1}^{H} \textit{R}_{i}(\mathbf{x}_t,\mathbf{a}_t)] 
\end{equation}

\begin{algorithm}
	\caption{MAML for Reinforcement Learning} 
	\label{alg:maml_alg}
	\begin{algorithmic}[1]
	\Require $p$($\mathcal{T}$): distribution over tasks 
	\Require $\alpha, \beta$: step size hyperparameters
	    \State randomly initialize $\theta$
	    \While{not done}
	        \State Sample batch of tasks $\mathcal{T}_{i} \sim p$($\mathcal{T}$)
	        \For {all $\mathcal{T}_{i}$}
	            \State Sample $K$ trajectories $\mathcal{D} = \{\mathbf{x_1, a_1, ..., x_H, a_H}\}$ 
	            \Statex \hspace{1cm}using $f_\theta$ in $\mathcal{T}_{i}$
	            \State Evaluate $\nabla_{\theta}\mathcal{L}_{\mathcal{T}_i}(f_\theta)$ using $\mathcal{D}$ and $\mathcal{L}_{\mathcal{T}_i}$ in Equation 
	            \Statex \hspace{1.1cm}3. 
	            \State Compute adapted parameters with gradient 
	            \Statex \hspace{1.1cm}descent: 
	            \Statex \hspace{1.1cm}$ \theta_{i}' = \theta - \alpha\nabla_{\theta}\mathcal{L}_{\mathcal{T}_i}(f_\theta)$
	            \State Sample trajectories $\mathcal{D}_i'= \{\mathbf{x_1, a_1, ..., x_H, a_H}\}$ 
	            \Statex \hspace{0.9cm} using  ${f}_{\theta_i}'$ in ${\mathcal{T}}_{i}$
	        \EndFor
	        \State Update $\theta = \theta - \beta\sum_{\mathcal{T}_{i} \sim p(\mathcal{T})} \mathcal{L}_{\mathcal{T}_i}(f_{\theta_{i}'})$  using each $\mathcal{D}_i'$ 
	        \Statex \hspace{0.5 cm} and $\mathcal{L}_{\mathcal{T}_i}$ in Equation 3.
	    \EndWhile
	\end{algorithmic} 
\end{algorithm}

\subsection{TD3}
Twin Delayed Deep Deterministic Policy Gradient (TD3) method proposed by \cite{fujimoto2018addressing} aims to reduce the overestimation bias seen in DDPG methods \cite{ddpg} and Deep Q learning based methods  such as \cite{dql}. These methods use a target network which is known to provide stability in deep reinforcement learning. However Fujimoto et al. in \cite{fujimoto2018addressing} noted that even when using a target network to update the policy network, the use of fast updating target networks still results in divergent behaviors. To mitigate this issue, Fujimoto et al. proposed to delay the policy network updates until the target network's error is as small as possible. This is achieved by only updating the policy after the target network is updated after a fixed $d$ iterations. To ensure that the error is small, the target network is updated in the following manner: 
\begin{equation}
 \theta' = \tau{\theta} + (1-\tau){\theta'}
\end{equation}
$\theta'$ corresponds to the target network parameter, $\theta$ corresponds to the the policy network parameters, and $\tau$ corresponds to the soft target update parameter which is initialized to a value between 0 and 1.  Algorithm \ref{alg:td3_alg} presents the overall TD3 algorithm used to train the baseline agent.

\begin{algorithm}
	\caption{TD3 Algorithm} 
	\label{alg:td3_alg}
	\begin{algorithmic}[1]
	\State Initialize critic networks $Q_{\theta_{1}}$,$Q_{\theta_{2}}$, and actor network $\pi_{\phi}$ with random parameters $\theta_{1}$,  $\theta_{2}$,  $\phi$ 
	\State Initialize target networks $\theta_{1}' \leftarrow \theta_{1}$, $\theta_{2}' \leftarrow \theta_{2}$, $\phi' \leftarrow \phi$
	\State Initialize replay buffer $\mathcal{B}$
	\For{$\mathbf{t}=1$ to $T$}
	    \State Select action with exploration noise $a \sim \pi_{\phi}(s) + \epsilon$, 
	    \Statex \hspace{0.6cm}$\epsilon \sim \mathcal{N}(0, \sigma)$ and
	    observe reward $r$ and new state $s'$
	    \State Store ($s, a, r, s'$) in $\mathcal{B}$
	    
	    \State Sample mini-batch of $N$ transitions  ($s, a, r, s'$) from $\mathcal{B}$ 
	    \Statex \hspace{0.4cm} and compute: 
	    \Statex \hspace{0.6cm}$\widetilde{a} \leftarrow \pi_{\phi'}(s') + \epsilon $  
	    \Statex \hspace{0.6cm}$\epsilon \sim clip(\mathcal{N}(0, \sigma), -c, c)$
	    \Statex \hspace{0.6cm}$y = \gamma min_{i=1,2} Q_{\theta_{i}'}(s',\widetilde{a})$
	    \State Update critics ${\theta}_{i}$ $\leftarrow$ $\argmin_{\theta_{i}}$ $\frac{1}{N}$ $\sum (y - 
	    Q_{\theta_{i}}(s,a))^2$
	    \State \textbf{if} $t$ mod $d$ \textbf{then}
	    \State \hspace{0.6cm} Update $\phi$ by the deterministic policy gradient: 
	    \Statex \hspace{1.1cm} $\nabla_{\phi}J(\phi) = \frac{1}{N} \sum \nabla_{a}Q_{\theta_{1}}(s, a) |_{a = \pi_{\phi}(s)} \nabla_{\phi}\pi_{\phi}(s)$
	    \State \hspace{0.6cm} Update target networks: 
	    \Statex \hspace{1.2cm} $\theta_{i}' \leftarrow \tau \theta_{i} + (1 - \tau)\theta_{i}'$
	    \Statex \hspace{1.2cm} $\phi' \leftarrow \tau \phi + (1 - \tau)\phi'$
	\EndFor
	\end{algorithmic} 
\end{algorithm}
\section{METHODS}

\subsection{MAML experimental setup}
Each experiment was run for at most 200 meta-iterations. Within each meta-iteration, we trained on 20 different tasks (i.e. 20 different uniquely generated environments, generated pseuodo-randomly). For each task, we generated 20 trajectories where each trajectory could be at most 300 steps.  We used a stochastic policy network to solve this problem, which is represented as a simple 2 layer feed-forward neural network with 100 hidden nodes at each layer, a ReLU activation for hidden layer, and a tanh activation function at final layer for normalizing actions. Furthermore when computing the network parameters, we do not use any first-order approximations as discussed in \cite{maml}.

\subsection{TD3 experimental setup}
For TD3, we used a deterministic policy network represented as a 2 layer feed-forward neural network with 256 hidden nodes at each layer, with ReLU activation for hidden layer, and a tanh activation function at final layer for normalizing actions. The target value networks $Q_1$ and $Q_2$ have the same architecture as that of the policy network. We only output a single value and do not use any activation functions at the last layer, however. We vary the replay buffer size between 20,000 and 100,000 and fill it with either 10,000 or 90,000 samples. At most, we collect a total of 1,000,000 samples. The smoothing parameter $\tau = 0.005$ and the policy network is updated after 2 updates of the target value networks.

\subsection{State and Action representation}
The environment we use functions with a 135-dimensional state space. 129 of these data points are LIDAR data which represents the robot's field of vision 90 degrees to its left and right. Six data points represent the current location of the robot and target location which are represented by x, y, and angle coordinates. This environment takes a 2-dimensional action vector that indicates how much thrust should be applied to the robot and at what angle.

\subsection{Reward Functions}
For both TD3 and MAML, we experimented with two different reward functions. The primary reward function of the environment was the distance to the goal, and thus did not encourage the agent to find an optimal path around any obstacles. Thus, we construct the following reward functions because this original reward signal is not conducive to learning a task through simulation: 

The first reward function we crafted, motivated by work done by \cite{yang_zhang}, is described below:
\begin{equation}
    R_{1} = r_{target} + r_{obstacle} 
\end{equation}

\begin{align*}
     r_{target} = \left\{
            \begin{array}{ll}
                -0.2 * dist_{target} & \quad \textnormal{goal not reached} \\
                -0.2 * dist_{target} + 200 & \quad \textnormal{goal reached}
            \end{array}
        \right.      
\end{align*}

\begin{align*}
r_{obstacle} = \left\{
        \begin{array}{ll}
            -1.5 & \quad dist_{obs} < 0.25 \\
            -0.2(0.75-dist_{obs}) & \quad 0.25 \leq dist_{obs} < 0.75 \\
            0 & \quad dist_{obs} > 0.75
        \end{array}
    \right.    
\end{align*}

\noindent $dist_{target}$ and $dist_{obs}$ corresponds to the spatial distance between the robot and the target location and spatial distance between robot and obstacle respectively. 

This reward function was designed to create a continuous reward space. If the robot enters the critical zone under 0.25 units, then a lower negative cost is accumulated. If the robot is outside the critical zone but not in the safety zone (between 0.25 and 0.75 units), then the agent incurs a smaller negative cost – it grows larger as the robot moves closer to the safe zone. In the safe zone, no cost is incurred and the only penalty it incurs is from $r_{target}$. At any given step, the robot also receives a larger reward as it moves closer to the target and receives a large positive reward (+200) if it reaches the goal location.

The second reward function we crafted, can be found below:
\begin{equation}
    R_2 = \left\{
            \begin{array}{ll}
                200 & \quad \textnormal{robot reached goal} \\
                -100 & \quad \textnormal{robot collides with obstacle} \\
                -1 & \quad \textnormal{otherwise}
            \end{array}
        \right.    
\end{equation}

The second reward function was designed to remove any local maxima in the reward space – every step taken gives the robot a uniform cost (-1). Any obstacle collision deals a large negative reward (-100) and reaching the obstacle rewards the agent with a large positive reward (+200). The intuition behind this formulation was that the agent could potentially reach the global maximum solution easier than if it were in a continuous reward space.

\section{RESULTS}
\subsection{MAML experiment results}
We ran three different experiments for MAML. Experiment 1 ran with reward function $R_1$ but excluding the +200 reward if the agent reached the target. Through all 20 tasks used for each meta-iteration, the robot initially starts at a random position. The average returns for this experiment are plotted below in figure \ref{fig:exp1}:

\begin{figure}[h]
    \centering
    \includegraphics[scale=0.45]{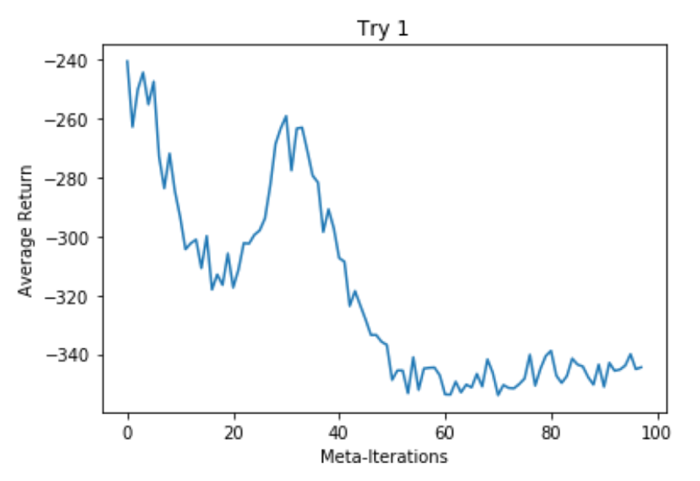}
    \caption{Average returns per meta iteration in MAML Experiment 1}  
    \label{fig:exp1}
\end{figure}

Experiment 2 was run with $R_{1}$ but with the +200 reward if the agent reached the target. We also created larger critical and safety zones, so the robot has time to react to negative costs if it enters into the critical zone. The average returns for this experiment are plotted below in figure \ref{fig:exp2}:

\begin{figure}[h]
    \centering
    \includegraphics[scale=0.45]{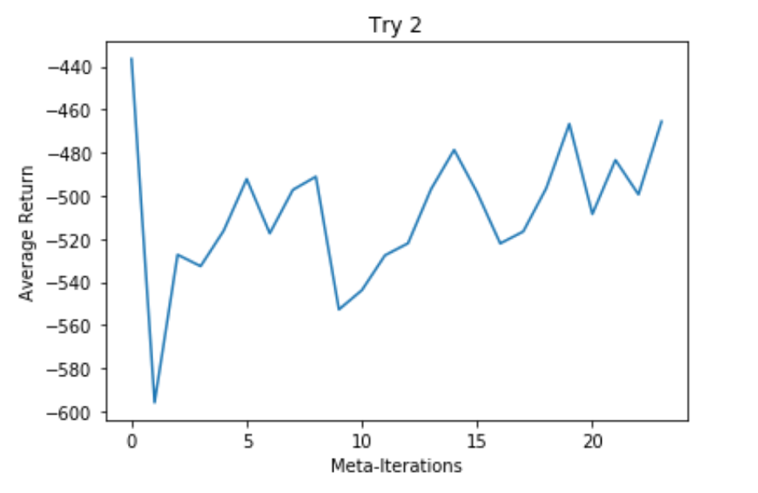}
    \caption{Average returns per meta iteration in MAML Experiment 2}  
    \label{fig:exp2}
\end{figure}

Experiment 3 was run with reward function $R_2$. In this experiment, each meta-iteration processed 20 tasks. For each task, the robot started at the same position while the obstacle and target locations changed. The average returns for this experiment are plotted below in figure \ref{fig:exp3}:

\begin{figure}[h]
    \centering
    \includegraphics[scale=0.45]{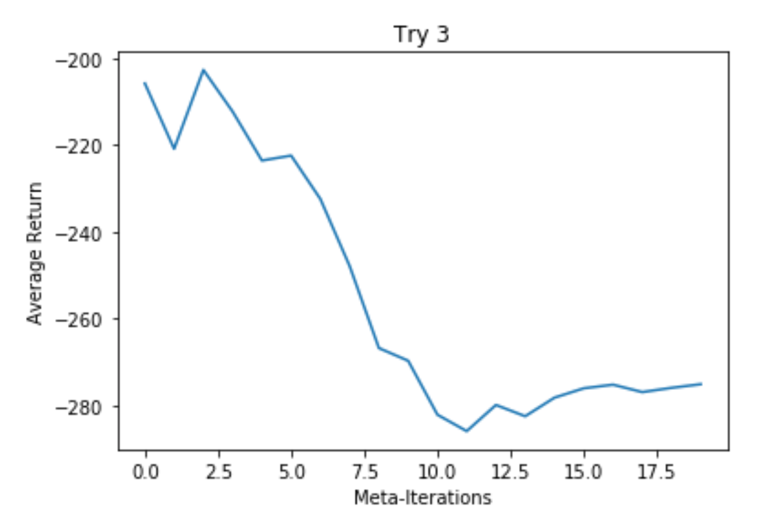}
    \caption{Average returns per meta iteration in MAML Experiment 3}  
    \label{fig:exp3}
\end{figure}

\subsection{TD3 experiment results}

We ran three different experiments with the TD3 algorithm. In Experiment 1, the agent was trained with TD3 with a replay buffer size of 20,000 and initialized with 10,000 samples. Furthermore TD3 collected a total of 100,000 samples and used the reward function $R_1$. The average return per episode is plotted below in figure \ref{fig:exp1_td3}:

\begin{figure}[h]
    \centering
    \includegraphics[scale=0.25]{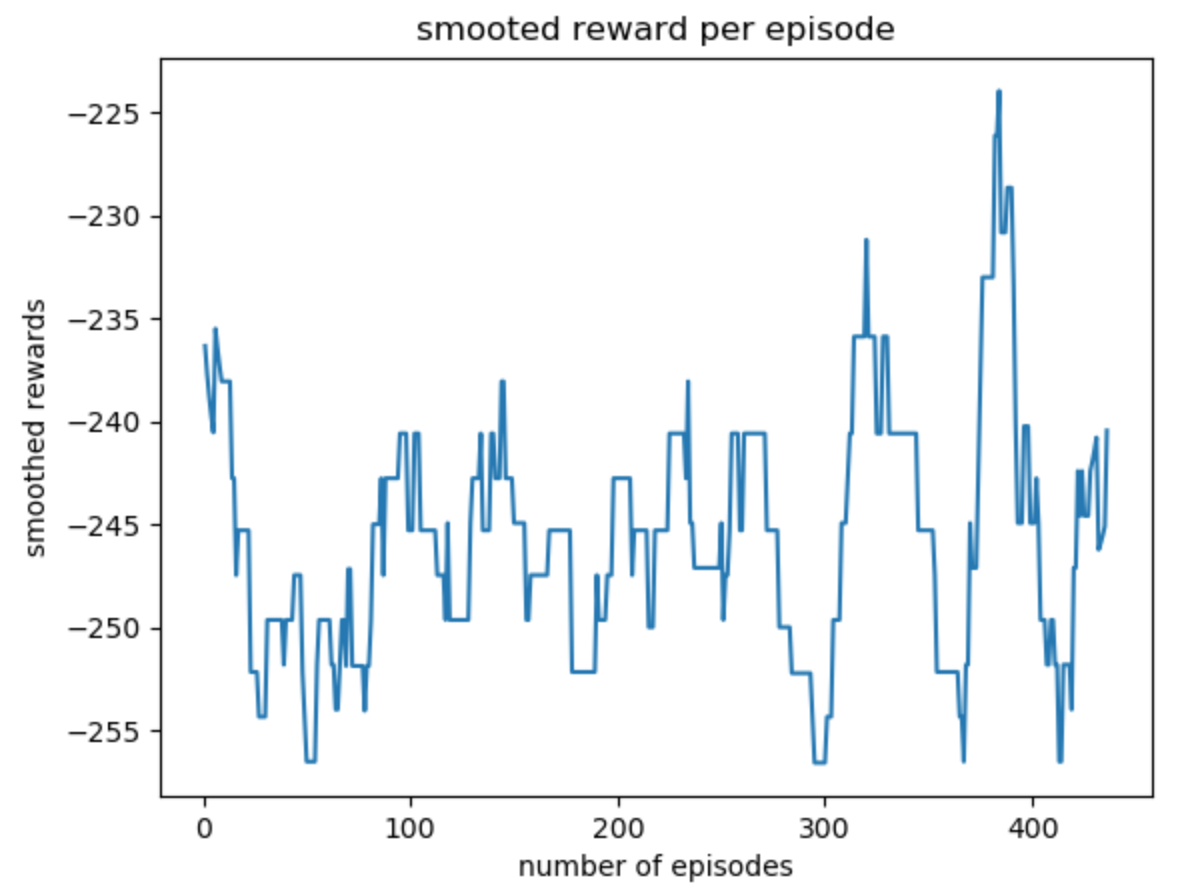}
    \caption{Average returns per episode in TD3 Experiment 1.}  
    \label{fig:exp1_td3}
\end{figure}

In Experiment 2, we used the experimental setup from Experiment 1 but with reward function $R_2$. The average return per episode is plotted below in figure \ref{fig:exp2_td3}:

\begin{figure}[h]
    \centering
    \includegraphics[scale=0.35]{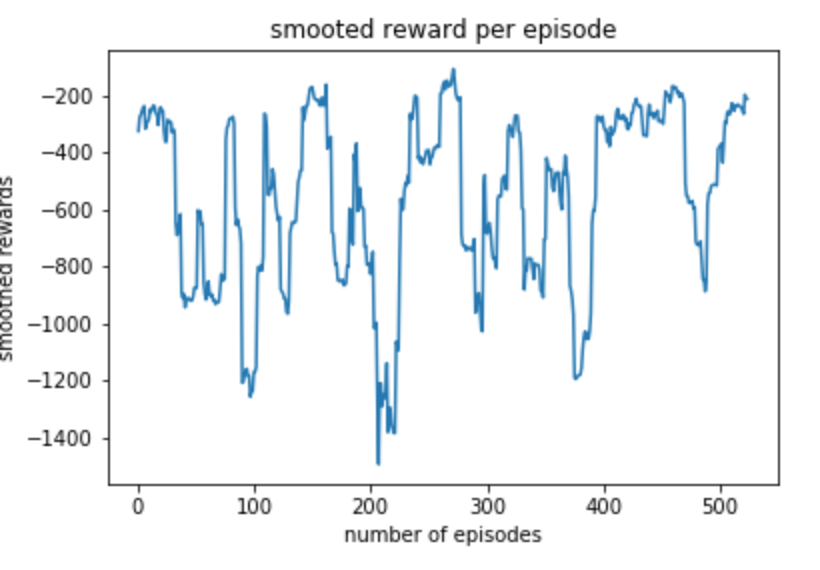}
    \caption{Average returns per episode in TD3 Experiment 2.}  
    \label{fig:exp2_td3}
\end{figure}

In Experiment 3, we used the same experimental setup from Experiment 1. This time, however, we collected a total of 1,000,000 samples and used a replay buffer of size 100,000 and initialized it with 90,000 samples. The average return per episode is plotted below in figure \ref{fig:exp3_td3}:

\begin{figure}[h]
    \centering
    \includegraphics[scale=0.35]{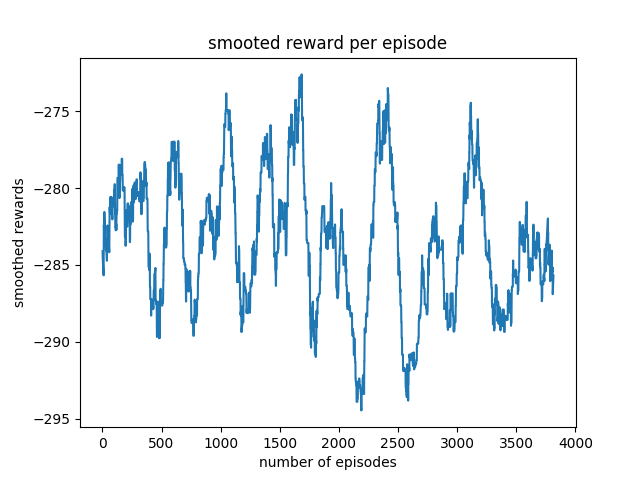}
    \caption{Average returns per episode in TD3 Experiment 3.}  
    \label{fig:exp3_td3}
\end{figure}

\section{DISCUSSION}

In the above experiments, the agent's average return either fluctuates or gets worse over time. Upon inspection of the first environment, we found that the robot never once reached the target. Indeed, the trajectories generated by the adapted model parameters result in the agent crashing into obstacles (which ends the simulation) or failing to reach the target within the designated time horizon limit. Thus, the trained policy found that crashing into an obstacle as quickly as possible would yield the maximum reward through all simulated episodes. For this reason, the average return per meta-iteration continues to decrease through training as shown in figure \ref{fig:exp1}.

In the second MAML experiment, we hypothesized that $R_{1}$ would properly backpropagate higher rewards in good episodes. Although the trained agent was slightly better than Experiment 1’s agent, the robot still did not reach the target through its simulations. We found that the robot terminated after angling itself in the direction of goal location, possibly because a local maximum reward had been reached. Through the meta-iterations we trained, the average return does increase but test simulations show that the robot does not advance far from its starting location. 

In third MAML experiment, $R_2$ was used to eliminate the possibility of falling into local maxima in the reward space. Additionally, to relax the task complexity, the agent was trained with environments where the robot starts at roughly the same position every time. However, we found that the robot still attempts to collide with the obstacle despite the strong intuition behind $R_2$. In this case, we believe that the agent needs to be trained for a longer time, possibly with a modified reward function that incorporates features from both $R_2$ and $R_1$. Furthermore, we believe that it is necessary to train the policy in a more hierarchical fashion - the agent should be trained on reaching the goal location initially before gradually introducing obstacles.

Throughout these experiments a common theme we observed is that the agent gets stuck at a local maximum and instead learns a reward-hacking based policy. The policy learned by the agent using TD3 further corroborates this notion. In both the first and second TD3 experiment, the reward function fluctuates drastically, which is due to changes in the environment. However, even with the fluctuation, the final learned policy remains the same.

During testing, we observed the agent would spin in circles and crash quickly into the obstacle, or spin until termination without reaching the goal. We believe that this behavior of reward hacking may be partly due to the replay buffer being quickly filled with samples from the learned policy, which we previously conclude is stuck in some local minima.

Furthermore, we believed that increasing the replay buffer and collecting more samples would mitigate the reward hacking. Unfortunately, the behavior persisted as seen in the third TD3 experiment, where average returns still fluctuate in figure \ref{fig:exp3_td3}. One potential reason for this might be related to how the target value network is updated. For instance, when the model begins training, we observe that the target network's loss plummets to near zero once the replay buffer is full of sampled collected by the learned policy. Since the actor network uses the target network to update its parameters, the abnormally low loss indicates that the same policy is used regardless of the number of samples collected. These experiments show that even with a straightforward and intuitive reward function that produces a clear reward signal depending on the robot's location, the learned policy through MAML or TD3 does not yield a working policy. 

\section{CONCLUSION}
In this paper, we used MAML and TD3 to train a policy that navigates a robot to a goal while avoiding any obstacles. We observed that the learned policy utilizes reward hacking to achieve some local minima in cost, despite formulating a reward signal that produces a piece-wise smooth function for any given location in the environment. Primarily, we believe that the agent's failure can be attributed to our manually constructed reward functions and the algorithms' inability to learn from few successful cases generated during training simulations.

To further expand on these experiments, we can use MAML using hierarchical training to ensure that the agent masters the task in distinct stages. Another possible direction is to learn a better reward function through inverse reinforcement learning methods. Once we learn such a reward function, we could then apply MAML and TD3 to see whether we can achieve better performances. 

Lastly in order to reproduce the results of our paper, please find the code at https://github.com/abhi-iyer/meta-learning. 

\section*{ACKNOWLEDGMENT}

We would like to thank Jacob Johnson and Michael Yip for their guidance throughout this project. We also thank Tristan Deleu and Scott Fujimoto for their open source implementation of MAML \cite{deleu2018mamlrl} and TD3 \cite{fujimotoCode} respectively that was very valuable in this project. 

\bibliographystyle{IEEEtran}
\bibliography{references}

\end{document}